# A Delay Compensation Framework Based on Eye-Movement for Teleoperated Ground Vehicles

Qiang Zhang, Lingfang Yang, Zhi Huang, Xiaolin Song

***Abstract*—An eye-movement-based predicted trajectory guidance control (ePTGC) is proposed to mitigate the maneuverability degradation of a teleoperated ground vehicle caused by communication delays. Human sensitivity to delays is the main reason for the performance degradation of a ground vehicle teleoperation system. The proposed framework extracts human intention from eye-movement. Then, it combines it with contextual constraints to generate an intention-compliant guidance trajectory, which is then employed to control the vehicle directly. The advantage of this approach is that the teleoperator is removed from the direct control loop by using the generated trajectories to guide vehicle, thus reducing the adverse sensitivity to delay. The delay can be compensated as long as the prediction horizon exceeds the delay. A human-in-loop simulation platform is designed to evaluate the teleoperation performance of the proposed method at different delay levels. The results are analyzed by repeated measures ANOVA, which shows that the proposed method significantly improves maneuverability and cognitive burden at large delay levels (>200 ms). The overall performance is also much better than the PTGC which does not employ the eye-movement feature.**

## I. Introduction

With the advantages of safety, high efficiency, and low cost, teleoperated ground vehicles have been widely applied in the military and civilian fields[1]. In a teleoperated ground vehicle, the operator and the vehicle are spatially separated, and the operator's commands are transmitted to the vehicle via a wireless network.

The main challenge for teleoperation is that the inevitable communication network delays may lead to asynchrony between sending commands to vehicle and receiving responses from vehicle, resulting in the degradation of vehicle maneuverability. When the delay is slight, human operators can adapt themselves by predicting the response. However, when the round-trip delay is large, human's adaptability to delays degrades severely, leading to an aggravated cognitive workload and an unstable closed-loop system [2]. For example, human operators tend to oversteer due to a lack of a clear correspondence between their operations and vehicle responses, resulting in steering oscillations. Experiments show that the performance of teleoperation systems degrades when the video feedback delay is above 200 ms [3].

Researchers have proposed various delay compensation methods for teleoperation systems to address challenges caused by significant communication delays to improve system performance and maintain system stability. These approaches can be categorized into prediction-based and supervisory control-based methods[4].

Prediction-based methods, including the predictive display and predictive control, aim to compensate for delays by predicting vehicle motion or human operator commands to improve performance in continuous teleoperation. In the predictive display approach, the vehicle's dynamic response to the current operator's commands is predicted and immediately displayed to help the operator get timely feedback on their operations. Researchers focused[5] on state estimators to model the vehicle dynamics. With the help of predictive display, the operation efficiency is improved by 20%[6]. Prakash[7] proposed a method to provide human drivers with a predictive video stream, which was produced by combining the perspective projection with the corrections given by the Smith predictor in the control loop. Studies also use head-mounted VR devices to provide operators with more realistic 3D displays[8][9]. The predictive control approach predicts future control commands before they are sent to the remote vehicle to compensate for delays[10]. The Smith predictor[11] is one of the most commonly used methods to minimize the constant time delay in closed-loop systems. To deal with time-varying Internet delays, Thomas[12] designed an adaptive Smith predictor that employed a delay estimator based on the characteristic roots of delay differential equations to measure delay. The adaptive algorithm[13] was also studied to address time-varying delays and uncertainties in Internet-based teleoperation systems. The above predictive methods rely on vehicle or human models to make predictions. However, these models are not always readily available or feasible. Furthermore, modeling errors can degrade the performance of model-based solutions.

To overcome the disadvantages of the model-based method, the model-free prediction, which did not require accurate

(Corresponding author: Zhi Huang).

Qiang Zhang, Zhi Huang, Xiaolin Song, are with the School of Mechanical and Vehicle Engineering, Hunan University, Changsha 410000, China (e-mail: zhangqiang@hnu.edu.cn; huangzhi@hnu.edu.cn; jqysxl@hnu.edu.cn;).

Lingfang Yang is with the School of Civil Engineering, Hunan University, Changsha 410000, China (e-mail: yanglf@hnu.edu.cn)

knowledge of vehicle or human dynamics, was proposed, such as vehicle trajectory clothoid prediction based on the assumption of constant vehicle speed [14] and Kalman filter-based prediction using Taylor-series expansion [15]. A model-free predictor without assumptions on model behavior is further developed[4]. The predictor is a first-order time-delay system that is designed based on the stability of the coupling error and frequency-domain performance analysis. Thus, the proposed method is sensitive to the coupling error's time delay and frequency characteristics. Zheng[16] developed a blended architecture for vehicle heading prediction by combining the model-based approach's performance and the robustness of the model-free prediction. The performance of the blended architecture predictor is comprehensively analyzed by human-in-the-loop simulation experiments in [17]. Guo[18] extended the single-parameter model-free framework to a two-parameter version to enhance its tuning flexibility.

Supervisory control-based methods reduce the effects of overcorrected behavior on closed-loop systems by removing the operator from the control loop to minimize the control frequency. This method sacrifices efficiency for system stability. The move-and-wait is a simple supervisory control used to compensate for delays. However, this method is extremely inefficient. Zhu[19] utilized the constructed local occupancy grid map (OGM) to generate the guidance points for remotely operated vehicles and transformed the guidance point selection into trajectory selection. Zhang[20] presented a haptic-guided path-generation method. An operator draws the desired 2D path by drawing it in a large-scale haptic interface and then generates a traceable path based on a locally optimized path planner. These methods improve the vehicle movement speed when the move-and-wait strategy is employed. Since the trajectory generation still relies on manual selection, as the vehicle speed increases, it would fluctuate if the operator fails to select a correct trajectory timely. To address the high workload of the pick-to-go methods, Prakash[21][22] proposed a successive reference-poses-tracking (SRPT) method, which utilized a joystick steering wheel to generate successive reference poses and transmitted a reference trajectory to the vehicle in the form of reference poses. Schitz[23] proposed an interactive corridor-based path planning framework. A collision-free path is planned through manually specified corridors guiding to destination.

The environmental constraints and human cognitive ability may contribute to an automatic guidance trajectory generation. Nevertheless, they have not been fully explored in the current research. In our previous work[3], a PTGC framework for delay compensation is proposed, which uses the operator's historical commands and the LiDAR 3D point cloud to predict the operator's intended trajectory to guide the vehicle. Since the trajectory is generated with historical commands, it results in a lag in the predicted trajectories. Moreover, over-reliance on steering maneuvers to generate future trajectory does not help to relieve the operator's workload. Therefore, if the intention can be acquired before an operation, it may benefit the generation of guidance trajectories and the compensation for delays. From a human perspective, experienced drivers are accustomed to natural driving, i.e., gazing at the area of interest, steering with the steering wheel, and accelerating/decelerating with the pedals. Therefore, the driver's eye-movement behaviors imply intentions that override the operation[24]. Consequently, it would further enhance the performance of the PTGC framework when early intentions are captured from eye-movement, and human cognitive abilities are employed to process complex environments and provide more accurate and timely predictions.

This paper proposes a predicted trajectory guidance control based on the teleoperator's eye-movement to compensate for communication delays in ground vehicle teleoperation systems. The proposed method extracts human intentions, a high-level representation of human cognitive abilities, from eye-movement and generates intention-compliant guidance trajectories used to guide vehicles for delay compensation.

The main contributions of this paper are summarized as follows:

1) A parallel structured trajectory prediction model integrating eye-movement-based intention and multimodal trajectory prediction is proposed, improving prediction accuracy by utilizing cognitive abilities mined from visual AOI.

2) A novel eye-movement-based predicted trajectory guidance control (ePTGC) framework is developed for communication delay compensation, improving the teleoperation system's maneuverability and alleviating operators' workload compared to off-the-shelf approaches.

The remainder of this paper is organized as follows. Section II describes the system structure of the ePTGC framework. Section III presents the parallel structure-based prediction model for the intended trajectory. Human-in-the-loop experiment, experimental results, and discussions are described in Section IV. Finally, section V makes conclusions.

## II. Eye-Movement-Based Predicted Trajectory Guidance Control Framework

We propose an eye-movement-based predicted trajectory guidance control (ePTGC) framework, as shown in Fig. 1. The framework captures the teleoperator's insightful perceptions of the external environment and future intentions from the teleoperator's eye-movement. Then, it predicts a trajectory that matches the driver's intentions. The intended trajectory is used to guide the vehicle. The ePTGC framework consists of two modules: trajectory prediction and trajectory tracking. The teleoperator's commands and preprocessed eye-movement data are fed to the trajectory prediction module and combined with environmental information (i.e., 3D point cloud) to predict the teleoperator's intention and intended trajectory. The 3D LiDAR point cloud implies drivable areas and contributes to generating collision-free trajectories. Eye-movement implies the teleoperator's perception of the environment and intentions. Unlike previous studies that categorize eye-movement into fixation, saccade, and smooth pursuit, and assume that human intention is embedded in fixation[25]. Our study finds that eye-movement behavior is highly dynamic during driving due to the frequent acquisition of traffic information in a wide field of view, such as glancing left and right at intersections. Thus, not only fixation but also saccade and smooth pursuit are relevant to driving intention. Since eye-movement and control commands are sampled at 90 Hz and 10 Hz, respectively, to align the inputs, we used a bivariate Gaussian distribution to fit the driver eye-movements to generate the visual AOI (Area of Interest, $\boldsymbol{p}_{AOI}(t) \sim \mathcal{N}(\mu_x^{(t)}, \mu_y^{(t)}, \sigma_x^{(t)}, \sigma_y^{(t)})$) at the rate of 10Hz. Here, $\mu_x^{(t)}$ and $\mu_y^{(t)}$ are coordinates of the visual AOI center at time *t*. $\sigma_x^{(t)}, \sigma_y^{(t)}$ are the standard deviation in the horizontal and vertical direction, respectively.

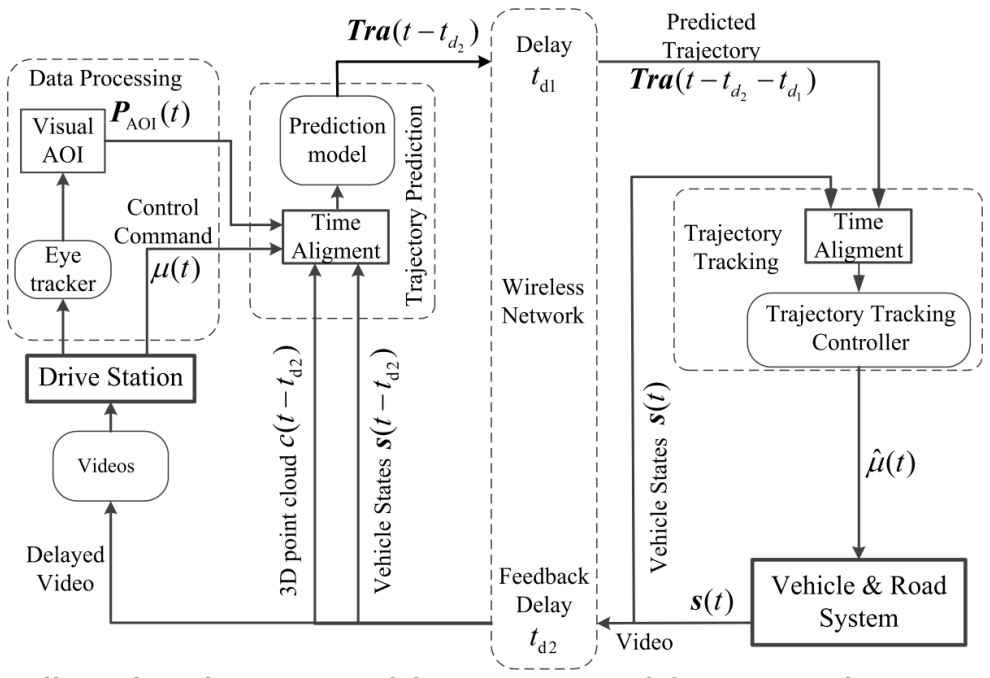

Fig. 1. Predicted trajectory guidance control framework

The teleoperator's control commands $\boldsymbol{\mu}(t)$ and visual AOI $\boldsymbol{p}_{\mathrm{AOI}}(t)$ correspond to the vehicle-road system states $\boldsymbol{s}(t - t_{\mathrm{d2}})$. Therefore, the predicted trajectory is denoted as $\boldsymbol{Tra}(t - t_{\mathrm{d2}})$ meaning prediction to inputs at time $t - t_{\mathrm{d2}}$. The predicted trajectory acts as the guidance trajectory and arrives at the vehicle after $t_{\mathrm{d1}}$. On the vehicle side, the predicted trajectory is denoted as $\boldsymbol{Tra}(t - t_{\mathrm{d}})$, here $t_{\mathrm{d}} = t_{\mathrm{d1}} + t_{\mathrm{d2}}$. The first $t_{\mathrm{d}}$ of the predicted trajectory is truncated and then aligned with $\boldsymbol{s}(t)$. The tracking controller outputs a steering command $\hat{\boldsymbol{\mu}}(t)$ according to the error between the actual and predicted trajectories. As long as the prediction horizon $T_{pre}$ is greater than the total delay $t_{\mathrm{d}}$, the delay can be compensated.

The advantage of this ePTGC framework is that, by deploying the tracking controller on the vehicle side and removing the teleoperator from the direct control loop, the proposed method is less sensitive to delays and vehicle dynamics, improves the stability of the trajectory following, and reduces the workload of the teleoperator.

## III. Intended Trajectory Prediction Based on Eye-Movement

To fully utilize the early driving intention inherent in eye-movement, we designed a parallel-structured trajectory prediction model, as shown in Fig. 2. The model consists of three parts: a driving intention (DI) module, a multimodal trajectory (MT) module, and a trajectory filtering (TF) module. The DI module predicts the driving intentions, and the MT module predicts the corresponding multimodal trajectories based on the driving intention. In the TF module, the trajectory-connected mode with the maximum probability is chosen as the predicted trajectory.

### *A. Trajectory prediction model*

As shown in Fig. 2, the DI and MT modules have the same encoding modules, including motion and context encoding. The motion encoding input features, including historical vehicle states $\boldsymbol{S}^{(t)}$, control commands $\boldsymbol{U}^{(t)}$ and visual AOI $\boldsymbol{P}_{AOI}^{(t)}$, are time-sequences. The LSTM has the advantage for time-sequence modeling. We employ LSTM (DI-LSTM and MT-LSTM) networks to encode the historical commands, vehicle states, and the visual AOI sequence. To obtain the road constraints, we convert the point cloud within a specified range into a binary image on a BEV grid[26]. Utilizing the powerful image processing capability of the Resnet network[27], we employ Resnet as a context encoder (DI-Resnet and MT-Resnet) to extract contextual features from the binary image of the 3D point cloud. A fully connected (FC) layer and SoftMax function are used as the DI decoder to output the probabilities of driving intention. LSTM acts as an MT decoder to generate multimodal trajectories corresponding to driving intentions.

At intersections, there are generally three maneuvers, i.e., going straight, right turn and left turn, denoted by 0, 1 and 2, respectively. We predict the trajectory of the future $T$ time steps with information in the past $T_h$ time steps. The historical vehicle states are denoted as

$$\boldsymbol{S}^{(t)} = \left[\boldsymbol{s}^{(t-T_h)}, \cdots, \boldsymbol{s}^{(t-i)}, \cdots, \boldsymbol{s}^{(t)}\right] \tag{1}$$

where $\boldsymbol{s}^{(t-i)} = [x_{t-i}, y_{t-i}, v_{t-i}, \theta_{t-i}]$ . $(x_{t-i}, y_{t-i})$ , $v_{t-i}$ and $\theta_{t-i}$ are the position, velocity, and heading at the time step $t - i, i \in (1, \cdots, T_{\mathrm{h}})$, respectively.

The historical control commands are denoted as

$$\boldsymbol{U}^{(t)} = \left[\boldsymbol{\mu}^{(t-T_h)}, \cdots, \boldsymbol{\mu}^{(t-i)}, \cdots, \boldsymbol{\mu}^{(t)}\right] \tag{2}$$

where $\boldsymbol{\mu}^{(t-i)} = [\delta_{t-i}, Th_{t-i}, Br_{t-i}]$. $\delta_{t-i}$, $Th_{t-i}$, and $Br_{t-i}$ are the steering, throttle, and brake commands at the time step $t - i, i \in (1, \cdots, T_{\mathrm{h}})$, respectively.

The historical visual AOI is denoted as

$$\boldsymbol{P}_{AOI}^{(t)} = \left[\boldsymbol{p}_{AOI}^{(t-T_h)}, \cdots, \boldsymbol{p}_{AOI}^{(t-i)}, \cdots, \boldsymbol{p}_{AOI}^{(t)}\right] \tag{3}$$

where $\boldsymbol{p}_{AOI}^{(t-i)} = \left[\mu_x^{(t)}, \mu_y^{(t)}, \sigma_x^{(t)}, \sigma_y^{(t)}\right]$.

Feeding the tensor $\mathbf{X}_t = cat\left(\boldsymbol{S}^{(t)}, \boldsymbol{U}^{(t)}, \boldsymbol{P}_{AOI}^{(t)}\right)$ into the motion encoders DI-LSTM and MT-LSTM, the maneuver features $\boldsymbol{M}_{\mathrm{DI}}^{(t)}$ and $\boldsymbol{M}_{\mathrm{MT}}^{(t)}$ are obtained, respectively.

The 3D point cloud in the range of $32\,\mathrm{m} \times 32\,\mathrm{m} \times 5\,\mathrm{m}$ ( length × width × height ) is converted to a binary image on the BEV grid. The grid has a resolution of 0.125m, and the resolution of the binary image is 256 × 256 pixels. Then the binary image is divided into two channels, one for the ground and the other for the non-ground, and a pseudo image is generated, denoted as $\boldsymbol{B}^{(t)}$. The context-constrained features $\boldsymbol{C}_{\mathrm{DI}}^{(t)}$ and $\boldsymbol{C}_{\mathrm{MT}}^{(t)}$ are obtained through DI-Resnet and MT-Resnet, respectively. $\boldsymbol{C}_{\mathrm{DI}}^{(t)}$ and $\boldsymbol{C}_{\mathrm{MT}}^{(t)}$ have a dimension of (512×1) and $\boldsymbol{M}_{\mathrm{DI}}^{(t)}$ and $\boldsymbol{M}_{\mathrm{MT}}^{(t)}$ have a dimension of (1×128). To fuse the extracted motion features with the contextual features and obtain more effective feature information, a multi-head attention mechanism is employed to obtain the fused features $\boldsymbol{A}_{\mathrm{DI}}^{(t)}$ and $\boldsymbol{A}_{\mathrm{MT}}^{(t)}$.

$$\boldsymbol{A}_{\mathrm{DI}}^{(t)} = Multihead\left(\boldsymbol{M}_{\mathrm{DI}}^{(t)}, \boldsymbol{C}_{\mathrm{DI}}^{(t)}, \boldsymbol{C}_{\mathrm{DI}}^{(t)}\right) \tag{4}$$

$$\boldsymbol{A}_{\mathrm{MT}}^{(t)} = Multihead\left(\boldsymbol{M}_{\mathrm{MT}}^{(t)}, \boldsymbol{C}_{\mathrm{MT}}^{(t)}, \boldsymbol{C}_{\mathrm{MT}}^{(t)}\right) \tag{5}$$

Feeding the tensor $\boldsymbol{A}_{\mathrm{DI}}^{(t)}$ into the DI Decoder to obtain the probability of the expected maneuver $\boldsymbol{P}^{(t)}(m_i|\boldsymbol{X})$:

$$\boldsymbol{P}^{(t)}(m_i|\boldsymbol{X}) = softmax\left(FC\left(c\boldsymbol{A}_{\mathrm{DI}}^{(t)}\right)\right) \tag{6}$$

where $m_i$ is the *i*-th maneuver, here $i \in (0, 1, 2)$, representing three maneuver modes. The sum of the probabilities of all three modes is one.

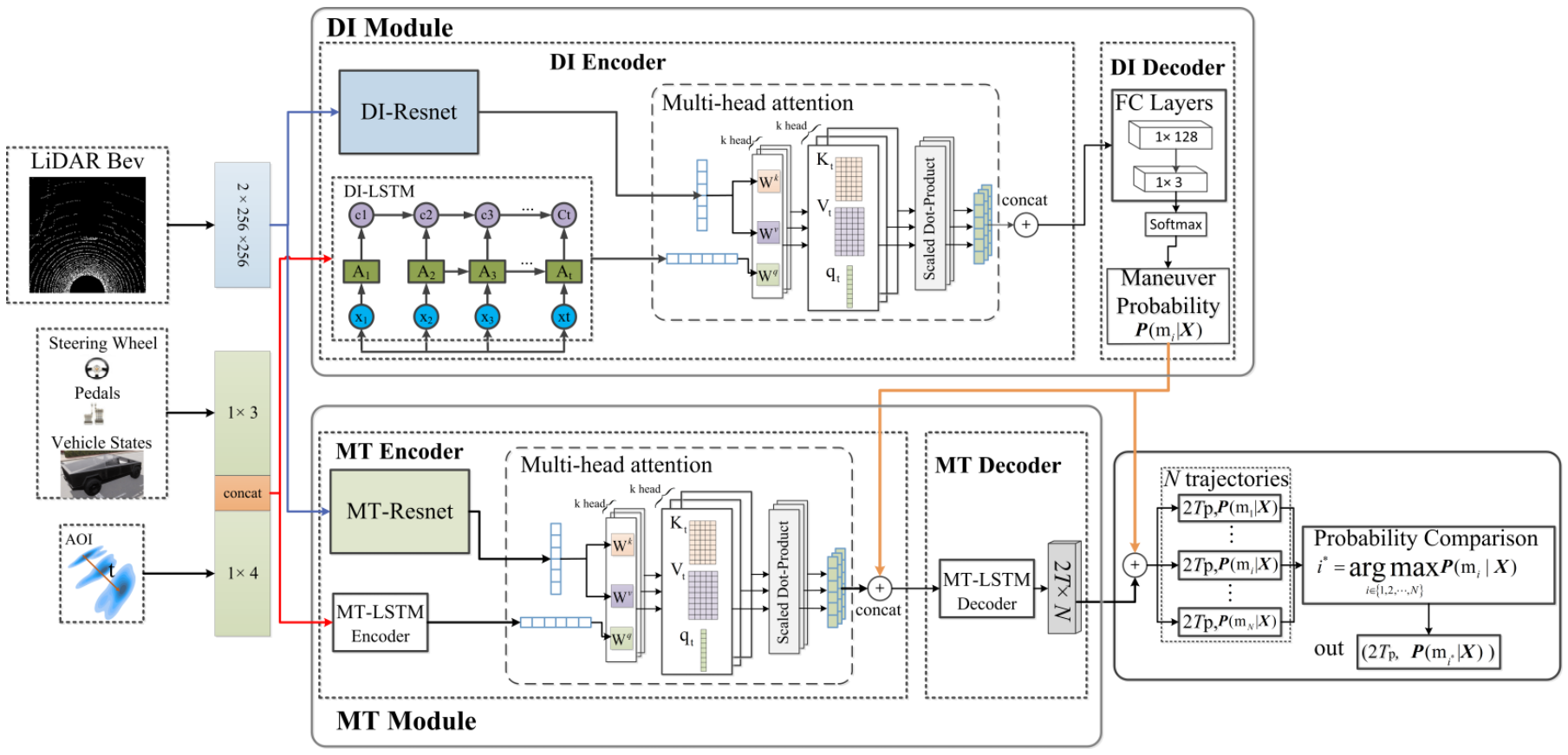

Fig. 2. Framework of visual AOI-based trajectory prediction

TABLE I
RESULTS OF DRIVING INTENTION PREDICTION

| Horizon | Models | Accuracy(%) | Going Ahead | | | Right Turns | | | Left Turns | | |
|---|---|---|---|---|---|---|---|---|---|---|---|
| | | | Pr (%) | Re (%) | F1 | Pr (%) | Re (%) | F1 | Pr (%) | Re (%) | F1 |
| 1s | E+C-DI | 95.19 | 95.56 | 95.90 | 95.73 | 92.55 | 93.83 | 93.19 | 96.23 | 94.58 | 95.40 |
| 2s | C-DI | 90.0 | 93.2 | 92.3 | 92.8 | 84.8 | 89.1 | 86.9 | 92.9 | 90.4 | 91.6 |
| | E-DI | 86.6 | 78.2 | 87.9 | 82.7 | 91.2 | 85.4 | 88.2 | 92.9 | 86.2 | 89.4 |
| | E+C-DI | 93.11 | 93.38 | 92.74 | 93.06 | 95.04 | 91.62 | 93.30 | 91.58 | 94.53 | 93.03 |
| 3s | E+C-DI | 88.17 | 82.13 | 88.06 | 84.99 | 94.88 | 84.92 | 89.63 | 87.31 | 90.51 | 88.88 |

$\boldsymbol{A}_{\mathrm{MT}}^{(t)}$ and $\boldsymbol{P}^{(t)}(m_i|\boldsymbol{X})$ are concatenated and fed into the MT decoder to obtain the multimodal trajectory $\boldsymbol{Tra}_{\mathrm{out}}^{(t)}$. Note that we are not to generate one trajectory but $N$ candidate trajectories, so the dimension of $\boldsymbol{Tra}_{\mathrm{out}}^{(t)}$ is $2T_{\mathrm{P}} \times N$, here $N = 3$ corresponding to the number of maneuver modes.

$$\boldsymbol{Tra}_{\mathrm{out}}^{(t)} = \left[\boldsymbol{Tra}_0^{(t)}, \dots, \boldsymbol{Tra}_i^{(t)}, \dots, \boldsymbol{Tra}_{N-1}^{(t)}\right] \quad (7)$$

where $\boldsymbol{Tra}_i^{(t)} = [(p_{x_i}^{(t+1)}, p_{y_i}^{(t+1)}), \dots, \left(p_{x_i}^{(t+j)}, p_{y_i}^{(t+j)}\right), \dots, (p_{x_i}^{(t+T)}, p_{y_i}^{(t+T)})]$, $\left(p_{x_i}^{(t+j)}, p_{y_i}^{(t+j)}\right)$ is the predicted waypoint of the $i$-th trajectory at the time step $t + j$, $i \in (0,1, \dots, N\text{-}1)$, $j \in (1, 2, \dots, T_{\mathrm{P}})$, and $N$=3. In the TF module, the trajectory corresponding to the maneuver mode with the maximum probability is selected as the output.

$$i^* = \underset{i \in \{0, \dots, N\text{-}1\}}{\operatorname{argmin}} \left(\boldsymbol{P}^{(t)}(m_i|\boldsymbol{X})\right) \quad (8)$$

The predicted trajectory is denoted by $\boldsymbol{Tra}_{i^*}^{(t)}$.

## B. Experiment Results of the DI Model

The dataset was constructed from driving simulators for training and evaluating predictive models. We applied the following model training settings: the past 20 steps (2s) for observation and the next 10, 20, and 30 steps (1s, 2s, and 3s) for prediction. Driving intentions are automatically labeled based on the deviation between the current and future heading, so we can get driving intentions for different time ranges. The collected data was segmented in a sliding window of 5 seconds, and a total of 24854 records were obtained. The ratio for training, validation, and testing is 3:1:1.

The DI model was trained and validated for different time ranges (1s, 2s, and 3s). An ablation experiment was conducted on the DI model to validate the significance of the input feature for a 2s prediction horizon. The Precision (Pr), Recall (Re), and F1 score(F1) are adopted for performance comparison. The experimental results are shown in Table I. Letters E and C represent visual AOI and context features, respectively. For example, E+C-DI denotes that visual AOI features and context features are employed.

According to the ablation experiment results, it can be concluded that the context feature and visual AOI are equally important for intention prediction, and the two features work synergistically to improve accuracy.

A longer intention prediction horizon enables the trajectory prediction model to better understand future trajectory modes. Nevertheless, the accuracy of intention prediction decreases significantly as the prediction horizon increases. Therefore, the E+C-DI pre-trained model with a 2s intention horizon is incorporated into the trajectory prediction module.

## C. Experiment Results of Trajectory Prediction

The results of ablation experiments and comparison experiments are presented in Table II. The following three trajectory prediction models are used for comparison. The MTP model is implemented with open-source code, and the other two models are implemented based on the literature.

- FF-LSTM[28]: A basic encoder-decoder based on Feed Forward LSTM, where the LSTM of the decoding module takes the output from the previous time step as the input to the next time step.
- MTP-LSTM[29]: A multimodal trajectory prediction model that employs the same input features as ours. The output tensor with the dimension of $N(2T_{\mathrm{p}} + 1)$ is decomposed into $N$ trajectories and their corresponding probabilities by the designed MTPloss module.
- TCN-MDN[30]: A model utilizes dilated temporal convolutional (TCN) and residual networks to construct codec networks and combines mixture density networks (MDN) to achieve multimodal trajectory prediction.

TABLE II
TRAJECTORY PREDICTION RESULTS

| Models | 1 seconds | | 2 seconds | | 3 seconds | |
|---|---|---|---|---|---|---|
| | FDE(m) | ADE(m) | FDE(m) | ADE(m) | FDE(m) | ADE(m) |
| FF-LSTM | 0.54 | 0.41 | 1.25 | 0.7 | 2.57 | 1.15 |
| MTP | 0.57 | 0.40 | 1.21 | 0.70 | 2.22 | 1.07 |
| TCN-MDN | 0.49 | **0.37** | 1.17 | 0.64 | 2.44 | 1.09 |
| E-model | 0.48 | **0.37** | 1.15 | 0.63 | 2.46 | 1.07 |
| C-model | 0.57 | 0.43 | 1.2 | 0.7 | 2.33 | 1.1 |
| E+C-model | **0.45** | **0.37** | **0.78** | **0.51** | **1.38** | **0.73** |

TABLE III
DISTRIBUTION OF LARGE FDE ERRORS FOR THE 3 SECONDS PREDICTION HORIZON

| Error | >1 m | >2 m | >3 m | >4 m |
|---|---|---|---|---|
| FF-LSTM | 31.3% | 20.3% | 9.5% | 5.0% |
| MTP | 37.5% | 18.6% | 7.6% | 2.8% |
| TCN-MDN | 33.5% | 16.0% | 7.4% | 4.7% |
| E+C-model | 30.9% | 8.9% | 2.6% | 1.1% |

The ADE and FDE are adopted as evaluation metrics, where the ADE is the average value of the L2 distance between ground truth and prediction results over the prediction horizon, and FDE is the average value of the L2 distance between ground truth and prediction results at the final time step $T_{\mathrm{P}}$. The prediction performance is evaluated for three prediction horizons: 1 second, 2 seconds, and 3 seconds. The results from the ablation experiments show that E+C-model performs the best, indicating that context feature and visual AOI are equally important for trajectory prediction.

Compared to other models, E+C-model achieves the best results for all prediction horizon cases. The E+C-model model and the TCN-MDN model show similar performance for the case of the 1s prediction horizon. As the predicted horizon increases, the accuracy of the E+C-model decreases more slightly than the other models. It is benefited from the pre-trained DI model that can accurately extract the driving intention to enhance the trajectory prediction. The results verify that the parallel prediction structure can improve the trajectory prediction accuracy. The FDE distributions of the 3s prediction horizon are also given in Table III to analyze the models' performance further. It is found that the errors of the E+C-model are mainly in the range of 0-2m(96.3%), and the proportion of large errors, e.g., >3m, is much smaller than that of other models, which further proves that the proposed model outperforms other models.

## IV. TELEOPERATION EXPERIMENTS AND ANALYSIS

### A. Human-in-loop simulation platform

A real-time driver-in-the-loop simulation platform has been developed for the teleoperation experiments, as shown in Fig. 3. The simulation platform consists of four sub-systems, i.e., the driving station, the vehicle-road system, the controller, and the communication network. At the driving station, a Logitech G27 joystick is used to generate steering, braking, and acceleration control commands, and a monitor is used for delayed visual feedback. A Tobii eye-tracker fixed to the monitor captures the real-time operator’s eye-movement.

The vehicle-road system is jointly constructed with Carla and Trucksim, in which Carla provides the scenario simulation and sensor signals, e.g., RGB front-view camera and 3D LIDAR, and Trucksim receives the control signals via CAN bus, simulates the vehicle dynamics and generates real-time vehicle states to update vehicle position in Carla.

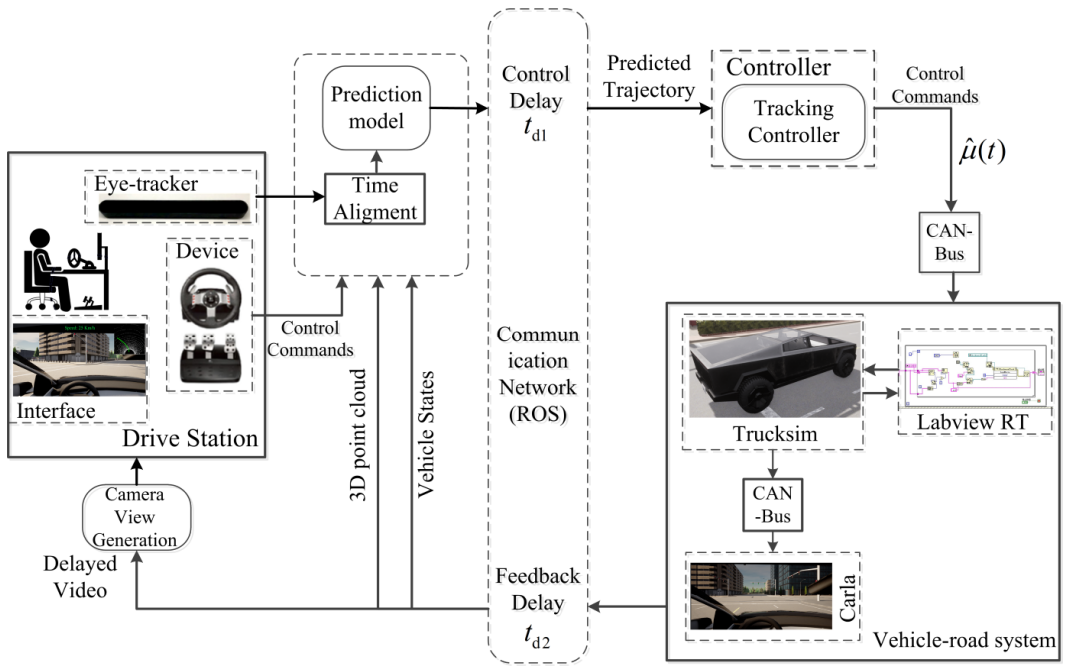

Fig. 3. Human-in-loop simulation for ground vehicle teleoperation system

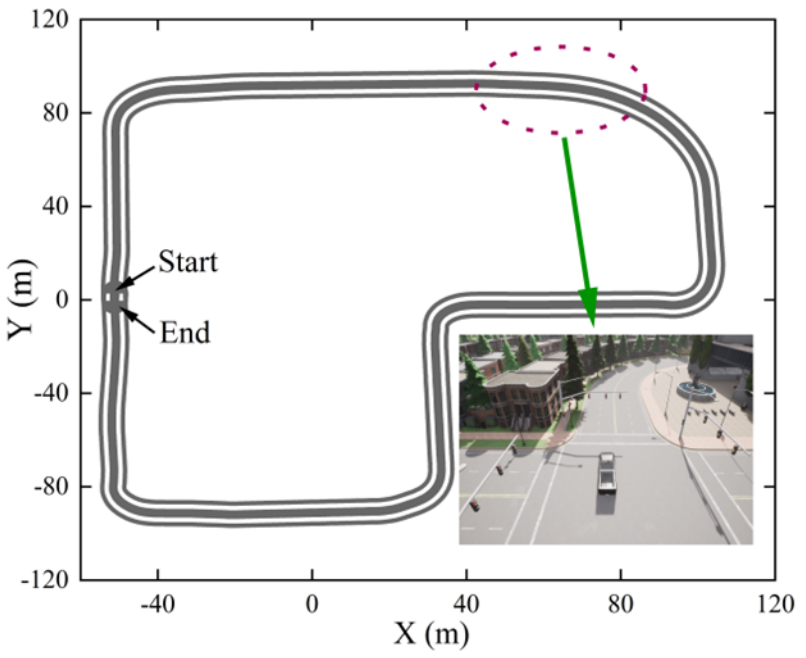

Fig. 4. Overview of the test route

A Stanley controller [3], is adopted for trajectory following, focusing only on the lateral control while using throttle commands directly from the remote manipulator for longitudinal control. This controller replaces the operator for lateral control but maintains the operator's longitudinal control over the vehicle, bolstering the operator's driving feeling.

The communication network is simulated by a first-in-first-out (FIFO) pipeline. A ROS node receives the video feedback and vehicle states in real-time. Then, the received message is pushed into the top of a queue, pulled out from the bottom, and sent to the driver station. Changing the queue length, the delay changes accordingly. Compared to a real teleoperation system, only the vehicle-road system dynamics and the communication system are simulated, while the human-machine interface is almost identical. Therefore, the simulation setup ensures the fidelity of the teleoperator's response to delays.

### B. Test Route and Experimental Procedures

The test route is designed to evaluate the performance of the proposed method at different delay levels. The route should include both straight and curved paths, where straight segments are used to evaluate the adjustment process and stability of path tracking. Curves are used to evaluate the transient response to step or slope inputs. Due to CARLA constraints, a structured roadway scenario was used. As shown in Fig. 4, beginning from the start point, the curved segment radii are 11m, 45m, 12m, 10m, 18m, and 19m. The total length of the route is 622m, including 442m straight segments and 180m curved segments.

TABLE IV
RM-ANOVA RESULT OF THREE METRICS

| Factor | log(*D2C*) | *TCT* | *SE* |
|---|---|---|---|
| Control | 0.000 | 0.000 | 0.000 |
| Delay | 0.000 | 0.000 | 0.000 |
| Control * Delay | 0.001 | 0.001 | 0.000 |

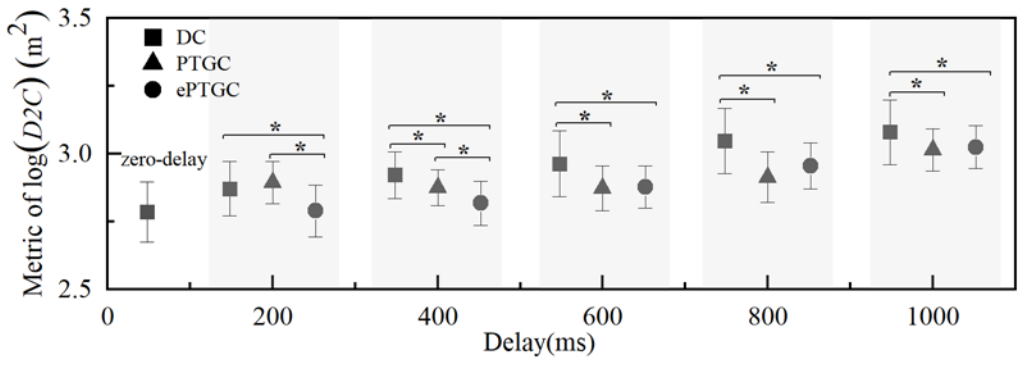

Fig. 5. Pairwise ANOVA comparison of log(*D2C*)

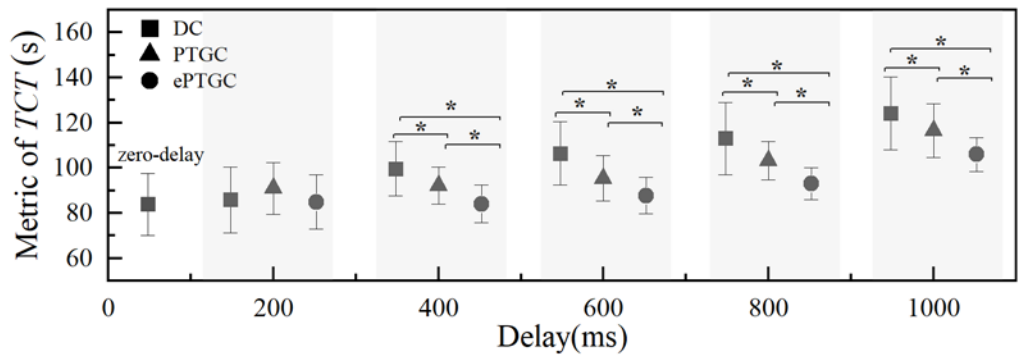

Fig. 6. Pairwise ANOVA comparison of *TCT*

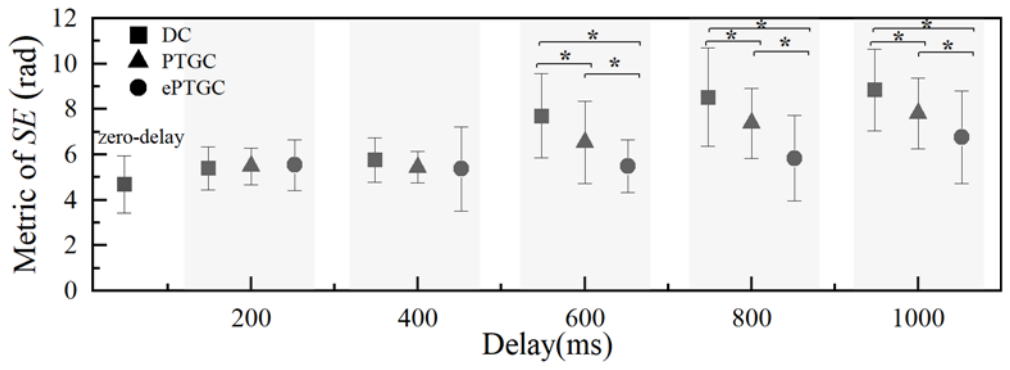

Fig. 7. Pairwise ANOVA comparison of *SE*

Three teleoperation methods were evaluated, including direct control (DC), PTGC, and ePTGC. DC refers to direct control by the operator's commands without any aids. PTGC is the predicted guidance control framework proposed in [3], which utilizes only driver operations and environmental constraints to predict the guidance trajectory. Experiments were performed at five delay levels from 200 ms to 1000 ms at 200 ms intervals.

The experiment adopted a 5 x 3 within-subjects factorial design to determine the effects of delay magnitude and control method on maneuverability and cognitive burden. Taking the zero-delay case as a baseline, a total of 16 driving tasks were carried out, and each task was repeated three times. The order of tasks is randomized to reduce the impact of learning on individual scenes or single trajectories.

A total of nine drivers aged 22 ± 3 years were recruited for the experiment. All participants have at least one year of driving experience and normal vision. Participants were trained before the formal start of the task. The training phase helped participants adapt to the simulation platform's operation and verbally informed them of the details of the test task, including the driving task performance goals, i.e., completion time, deviation error, and steering effort. Participants were asked to complete tasks as quickly as possible during the experiment, but they were not informed of the task performance priority, which depended entirely on their driving habits and adaptation process. Participants were required to complete sixteen randomly assigned driving tasks, and each task was repeated three times. It was valid if the following cases were not observed.

1) The vehicle runs off the road for 5 seconds.
2) Vehicle rollover.
3) The average speed is less than 18 km/h.

## C. Performance metrics and analysis methods

This experiment aims to explore the effectiveness of the proposed control framework on the maneuverability and teleoperator's cognitive burden when suffering from communication delays. Three independent parameters, i.e., deviation to centerline (*D2C*), task completion time (*TCT*), and steering effort (*SE*), are used as performance metrics. *D2C* is defined as the area between the actual path and the route's centerline, indicating the degree of deviation from the centerline. *TCT* is the time it takes for a participant to complete one loop of the driving task as quickly and smoothly as possible. These two metrics reflect the teleoperation system's longitudinal and lateral maneuverability, respectively. A small value of these two metrics means good maneuverability. *SE* is calculated based on the mean absolute steering angle, which indicates the operator's efforts to teleoperate a vehicle, and a small *SE* indicates a light cognitive burden on the operator.

A two-way RM-ANOVA is used to evaluate the effects of the two independent variables (i.e., Delay Level and Control Method) on *D2C*, *TCT*, and *SE*. There are five levels of Delay Level, i.e., {200ms, 400ms, 600ms, 800ms, 1000ms}, and three levels of Control Method, i.e., {DC, PTGC, ePTGC,}.

Two null hypotheses for each metric are tested using an *F*-test based on the type III sum of squares and 95% confidence level. Two null hypotheses are detailed as follows:

HC: No significant difference in performance metrics when using different control methods.

HD: There is no significant difference in performance metrics between different delay levels.

## D. Experimental results and discussion

Discarding invalid data, a total of 395 valid records were collected. Since the data for the metric *D2C* did not conform to the homogeneity of variance, *D2C* was log-transformed, and ANOVA was done with log(*D2C*). The two-way RM-ANOVA for the three metrics was analyzed individually using a general linear model, as detailed in Tables IV.

The *P* values of the delay level *F*-tests for all three metrics are much less than 0.05, indicating that the HD hypothesis is rejected at the 95% confidence level. Regarding the effect of the Control Method, the *P* values for log(*D2C*), *TCT*, and *SE* are also less than 0.05, indicating that the HC hypothesis is rejected at the 95% confidence level. The *P* values of the Control*Delay factors for the three metrics log(*D2C*), *TCT*, and *SE* were 0.001, 0.001, and 0.000, respectively, indicating the same significant interaction effect between the Control Method and the Delay Level. Pairwise ANOVA comparisons were performed further to explore the performance differences between control methods at different delay levels. The results are shown in Fig. 5-7, where '*' indicates pairs with statistically significant differences. As shown in Fig. 5, the ePTGC significantly improves *D2C* in all delay cases and outperforms the PTGC at small delay levels (≤400ms). As shown in Fig. 6, when the delay is ≤ 200ms, there is no significant difference in *TCT* among the three control methods. The reason might be that operators can adapt to the effect of the delay on speed at a small delay level. As the delay increases, the improvements of *TCT* of the ePTGC are significant, and ePTGC outperforms the PTGC methods. There is no significant difference in *SE* between the three control methods at delay levels ≤400ms. However, the difference is distinct at delay levels >400ms, where the *SE* of ePTGC is improved significantly, and the ePTGC method substantially outperforms the PTGC method.

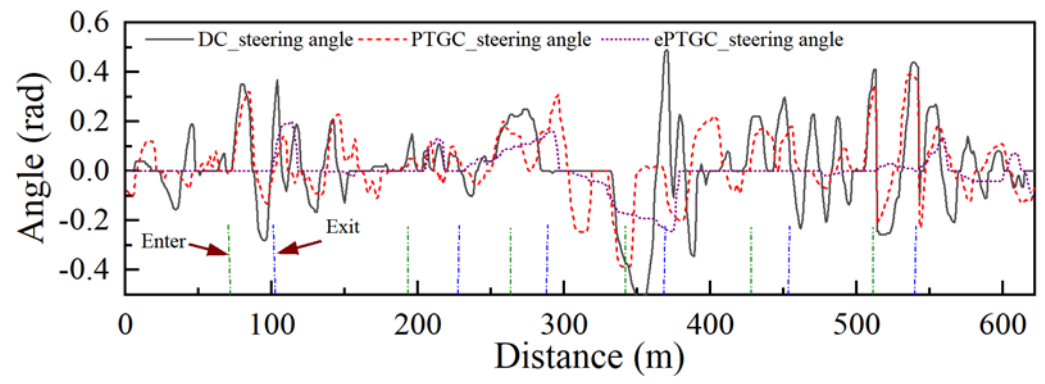

Fig. 8. Operator's steering command in the delay case of 800ms. The green and blue lines mark the spots entering and exiting turns

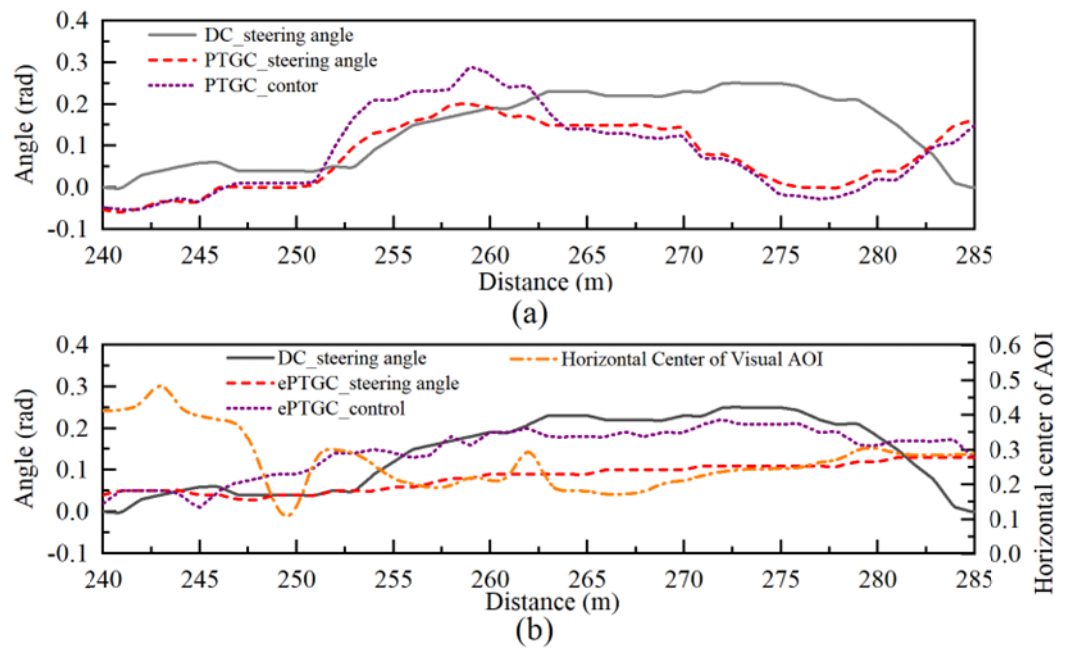

Fig. 9. Steering command of  the operator and Stanley controller at the third turn

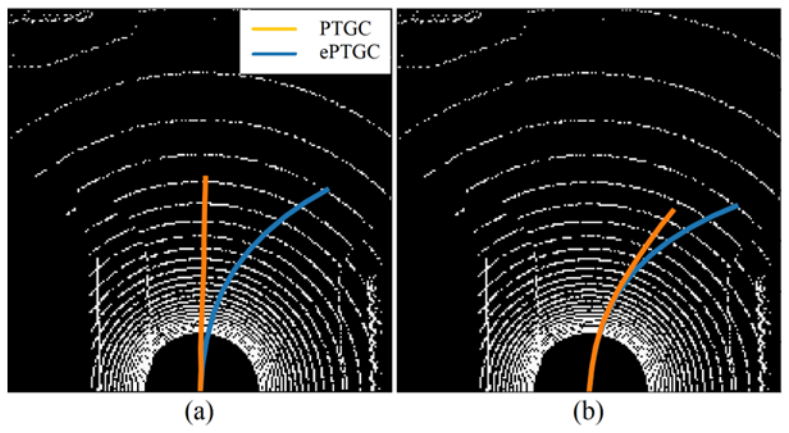

Fig. 10. Predicted trajectories at 250m and 255m, respectively.

According to the comprehensive analysis in Fig. 5-7, it can be found that although the ePTGC does not surpass the PTGC in *D2C* at large delay levels ($\geq 600ms$), its *TCT* and *SE* metrics are significantly improved and statistically significant. To further explore the rationale behind the considerable improvement in *SE* metrics, a case study (a delay of 800ms) is shown in Fig. 8 and Fig. 9. It can be found that the lack of synchronization between the control and the feedback caused the operator to oversteer during entering or exiting curved segments. The PTGC method mitigates the amplitude of oversteering by removing the teleoperator from the closed-loop control. However, since the predicted trajectories of PTGC are mainly extracted from operation commands, the effect of oversteering operation can be partially transferred to Stanley controller via the predicted trajectory. We can find that the controller's output varies with the steering wheel angle, as shown in Fig. 9(a). The ePTGC method fuses eye-movement

TABLE V

PERFORMANCE IMPROVEMENT VS. DELAY LEVELS

| Method | Delay | 400 ms | 600 ms | 800 ms | 1000 ms |
|---|---|---|---|---|---|
| PTGC | $P_{D2C}$ | 41% | 59% | 60% | 35% |
| | $P_{TCT}$ | 48% | 49% | 33% | 21% |
| | $P_{SE}$ | 0 | 38% | 28% | 23% |
| | $P_{ove}$ | **30%** | **49%** | **41%** | **27%** |
| ePTGC | $P^e_{D2C}$ | 82% | 57% | 44% | 27% |
| | $P^e_{TCT}$ | 98% | 83% | 68% | 45% |
| | $P^e_{SE}$ | 0 | 73% | 70% | 50% |
| | $P^e_{ove}$ | **60%** | **71%** | **61%** | **41%** |

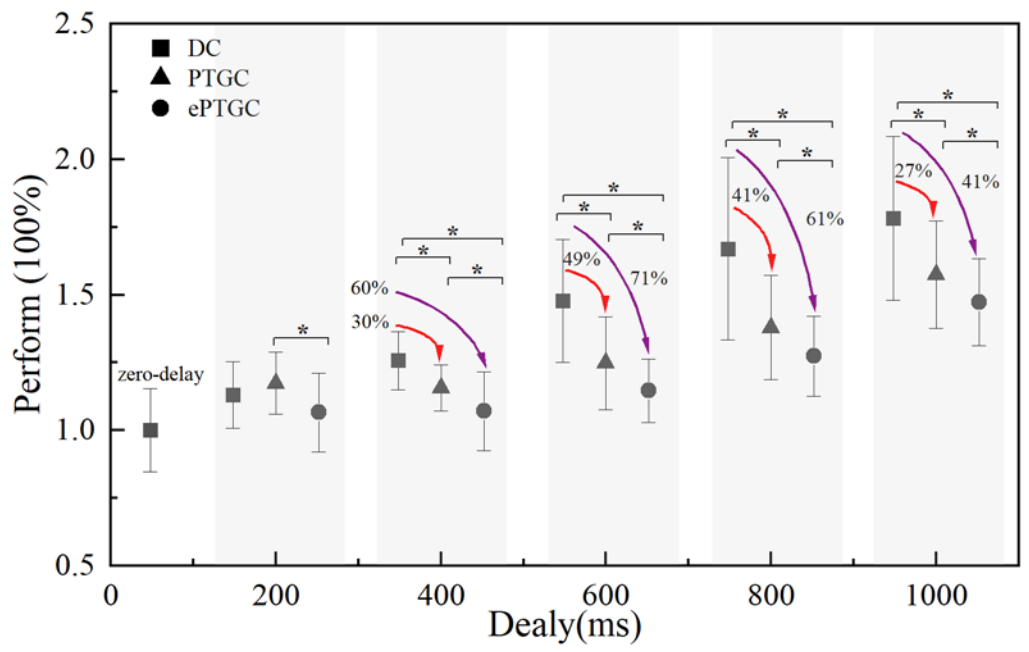

Fig. 11. Pairwise ANOVA comparison of the overall performance

and operation commands to extract the operator's intention and predict the guidance trajectory. During turning, the operator gazes at the destination by instinct. Therefore, the horizontal position of visual AOI would be distinct and steady. These characteristics of visual AOI contribute to a correct and continuous trajectory prediction, which helps reduce steering adjustment. In Fig. 9(b), it can be found that the control output of ePTGC varies with the visual AOI without excessive steering operation, which avoids misoperation due to operator misjudgment. The predicted trajectories of the PTGC and ePTGC methods before entering the third turn are shown in Fig. 10. Fig. 10(a) shows the predicted trajectories at 250m. As we can see from Fig. 9, there is no obvious steering operation at this point by all methods (including DC, PTGC, and ePTGC), resulting in the guidance trajectory predicted by the PTGC method is still a straight line. In contrast, the ePTGC method extracts the steering intention from the eye-movement, thus predicting a right-turn trajectory. However, until the operator's steering command is distinct, PTGC predicts the right turn, as shown in Fig. 10(b). This case demonstrates the benefits of fusing eye-movement in trajectory prediction.

Using the average performance metrics of the zero-delay case as the benchmark, the results were further normalized to analyze the overall performance improvement quantitatively. The normalized results of *D2C*, *TCT*, and *SE* metrics for the PTGC and ePTGC cases are denoted as $P_{D2C}$, $P_{TCT}$, $P_{SE}$ and $P^e_{D2C}$ , $P^e_{TCT}$ , $P^e_{SE}$, respectively. Taking $P^e_{D2C}$ for example, $P^e_{D2C}$ is calculated as

$$P^e_{D2C} = \frac{|r_c - r_d|}{|r_d - r_0|} \quad (9)$$

where $r_c, r_d$ *and* $r_0$ are the *D2C* means of ePTGC, DC and zero-delay cases, respectively.

Assuming that the three metrics contribute equally to the overall performance $P^e_{ove}$, we get

$$P^e_{ove} = \sigma^D * P^e_{D2C} + \sigma^T * P^e_{TCT} + \sigma^S * P^e_{SE} \quad (10)$$

where $\sigma^D = \sigma^T = \sigma^S = 1/3$.

Note that overall performance is derived based on the results of previous significance analyses. If the performance improvement is not statistically significant, set the coefficient $\sigma = 0$.

The normalized overall performance is shown in Table V, and the pairwise ANOVA comparison of the overall performance at different delay levels is shown in Fig.11. In small delay cases(≤200ms), neither the PTGC nor the ePTGC makes a significant improvement. It is because the operator can adapt to slight delay without any aids, so PTGC and ePTGC make no difference. As the delay increases, both PTGC and ePTGC significantly improve the overall performance. ePTGC surpasses the PTGC in terms of overall performance at all delay levels, which proves the feasibility of fusing eye-movement in trajectory prediction.

## V. Conclusion

This paper improved the predicted trajectory guidance control by fusing the teleoperator's eye-movement in the intended trajectory prediction to compensate for delays in teleoperation systems. The novelty is that the operation intention implied in the eye-movement is fully explored and utilized in the intended trajectory prediction, thus improving the accuracy of trajectory prediction. Ablation experiments and comparison experiments on intention and trajectory prediction have verified the feasibility and superiority of the proposed method. Human-in-the-loop simulation has been conducted to evaluate its application to ground vehicle teleoperation. Results show that the ePTGC significantly improves the maneuverability of the teleoperation system and the operator's cognitive burden when the delay is larger than 200 ms and achieves better *D2C, TCT, SE*, and overall performance than the PTGC method at almost all delay levels.

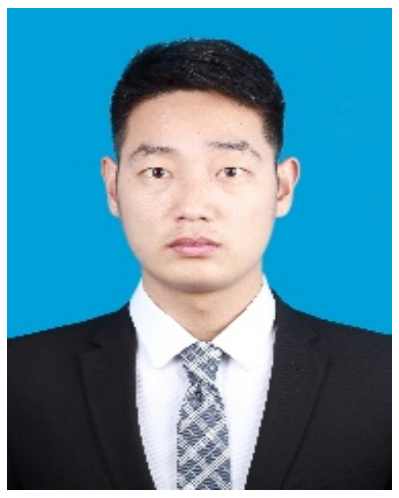

**Qiang Zhang** received the M. S. degree in mechanical design and theory from Zhengzhou University, Zhengzhou, China, in 2021. He is currently a doctoral student at the College of Mechanical and Vehicle Engineering, Hunan University, Changsha, China. His research interests include deep learning and collaborative control.

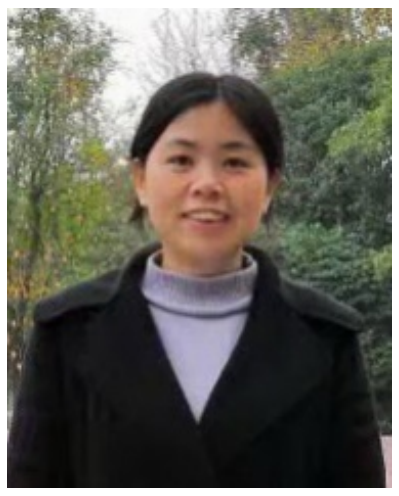

**Lingfang Yang** received the Ph.D. degree in civil engineering from Hunan University, Changsha, China. She is an associate professor with the College of Civil Engineering, Hunan University. Her research interests include intelligent transportation, deep learning and data mining.

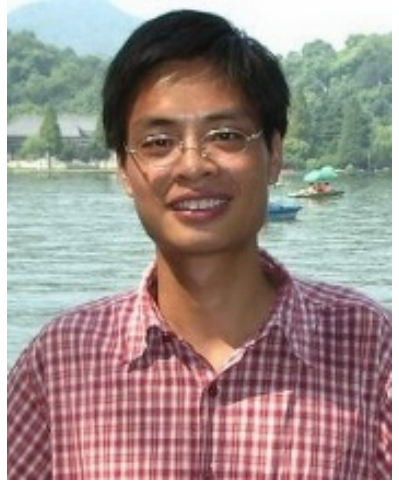

**Zhi Huang** received the Ph.D. degree in mechanical engineering from Hunan University, Changsha, China. He is an associate professor with the College of Mechanical and Vehicle Engineering, Hunan University. His research interests include advanced assistant driving, active safety, autonomous driving, embedded system and vehicle dynamics control.

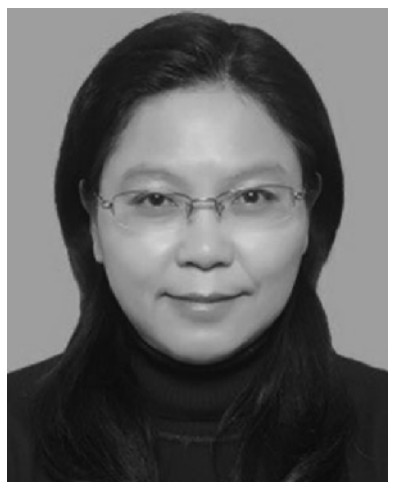

**Xiaolin Song** received the Ph.D. degree in mechanical engineering from Hunan University, Changsha, China. She is a professor with the College of Mechanical and Vehicle Engineering, Hunan University. Her research interests include active safety, vehicle dynamics simulation and control.